%% file: root.tex
\documentclass[letterpaper, 10 pt, conference]{ieeeconf}  

\input{preamble}

\begin{document}

\input{title}

\maketitle

\thispagestyle{empty}
\pagestyle{empty}
\input{commands}

\input{abstract}
\section{Introduction}
\label{sec:intro}

\input{intro}

\section{Related Work}
\label{sec:related}
\input{related}

\section{Problem Statement}
\label{sec:problem}
\input{problem}
\section{Learning from Inference-Time Execution}
\label{sec:pyplan}
\input{methodology}

\section{Experiments}
\label{sec:experiments}

\input{experiments}
\section{Conclusion}
\label{sec:conclusion}

\input{conclusion}






\section*{Acknowledgements}
The authors thank Catherine Glossop, Paul Zhou, and Kevin Black for their help and feedback. This work was supported in part by DARPA Contract FA8750-23-C-0080 (ANSR) and the DARPA Contract HR00112490425 (TIAMAT) and by Nissan and Toyota under the iCyPhy center.


\bibliographystyle{IEEEtran}
\bibliography{refs}

\appendix
\input{appendix}

\end{document}

%% file: preamble.tex
\usepackage{graphicx}
\usepackage{amsmath}
\usepackage{amssymb}
\usepackage{xcolor}
\usepackage{cprotect} 
\usepackage{tcolorbox}
\usepackage[colorlinks,allcolors=blue]{hyperref}
\usepackage{stfloats}
\usepackage{fancyvrb} 
\usepackage{fvextra}

%% file: title.tex
\IEEEoverridecommandlockouts                              

\overrideIEEEmargins                                      




\title{\LARGE \bf
Learning Affordances at Inference-Time for \\ Vision-Language-Action Models
}

\author{Ameesh Shah$^1$, William Chen$^1$, Adwait Godbole$^1$, Federico Mora$^1$, \\ Sanjit A. Seshia$^1$, Sergey Levine$^{1, 2}$\thanks{$^{1}$UC Berkeley $^{2}$ Physical Intelligence. Email: \href{mailto:ameesh@berkeley.edu}{ameesh@berkeley.edu}.} \\ \href{https://liten-vla.github.io/}{https://liten-vla.github.io/}}

%% file: commands.tex
\newcommand{\Reals}{\mathbb{R}}

\newcommand{\po}{{\preceq}}
\newcommand{\teacher}{\mathcal{T}}
\newcommand{\mOracle}{\mathcal{M}}
\newcommand{\cOracle}{\mathcal{C}}

\newcommand{\actions}{\mathcal{A}}
\newcommand{\observations}{\mathcal{O}}
\newcommand{\environ}{\mathcal{M}}
\newcommand{\alphabet}{\Sigma}
\newcommand{\transition}{\delta}
\newcommand{\initstate}{q_0}
\newcommand{\acceptstates}{F}
\newcommand{\vlm}{\pi^{\text{plan}}}
\newcommand{\method}{LITEN}
\newcommand{\pihigh}{\pi^\text{high}}
\newcommand{\pilow}{\pi^\text{low}}
\newcommand{\yes}{\succ}
\newcommand{\no}{\prec}
\newcommand{\inc}{\|}
\newcommand{\tasklanguage}{\ell}
\newcommand{\subtasklanguage}{\ell'}
\newcommand{\traj}{\tau}

\newcommand{\eqdef}{\mathrel{\stackrel{\makebox[0pt]{\mbox{\normalfont\tiny def}}}{=}}}

\newcommand{\Distr}[1]{\mathrm{Distr}(#1)}

\newcommand{\ameesh}[1]{\textcolor{violet}{AS: #1}}
\newcommand{\adwait}[1]{\textcolor{blue}{AG: #1}}
\newcommand{\sanjit}[1]{\textcolor{magenta}{SS: #1}}

\definecolor{bluegray}{rgb}{0.4, 0.6, 0.8}
\definecolor{lightgray}{rgb}{0.9, 0.9, 0.9}
\newcommand{\will}[1]{\textcolor{bluegray}{WC: #1}}
\newcommand{\sergey}[1]{\textcolor{red}{SL: #1}}

\newcommand{\todo}[1]{\textcolor{orange}{TODO: #1}}

\newtcolorbox{promptbox}{
  colback=gray!10,       
  colframe=white,        
  boxsep=5pt,            
  arc=0pt,               
  left=5pt, right=5pt,   
  top=5pt, bottom=5pt,   
  boxrule=0pt,           
  fontupper=\footnotesize\ttfamily    
}

\newenvironment{FullWidthPrompt}{
        \begin{tcolorbox}[
            width=\textwidth,
            colback=gray!10,
            colframe=white,
            boxsep=5pt,
            arc=0pt,
            left=5pt, right=5pt,
            top=5pt, bottom=5pt,
            boxrule=0pt,
            fontupper=\footnotesize\ttfamily
        ]
}{
        \end{tcolorbox}
}

\newtheorem{example}{Example}

%% file: abstract.tex
\begin{abstract}

Solving complex real-world control tasks often takes multiple tries: if we fail at first, we reflect on what went wrong, and change our strategy accordingly to avoid making the same mistake.
In robotics, \textit{Vision-Language-Action} models (VLAs) offer a promising path towards solving complex control tasks, but lack the ability to contextually and dynamically readjust behavior when they fail to accomplish a task. In this work, we introduce \textit{Learning from Inference-Time Execution} (\textbf{\method}), which connects a VLA low-level policy to a high-level VLM that conditions on past experiences by including them in-context, allowing it to learn the affordances and capabilities of the low-level VLA. 
Our approach iterates between a \textit{reasoning} phase that generates and executes plans for the low-level VLA, and an \textit{assessment} phase that reflects on the resulting execution and draws useful conclusions to be included in future reasoning contexts. Unlike similar approaches to self-refinement in non-robotics domains, \method\ must reflect on unstructured real-world robot trajectories (e.g., raw videos), which requires structured guiderails during assessment.
Our experimental results demonstrate \method\ is able to effectively learn from past experience to generate plans that use high-affordance instructions to accomplish long-horizon tasks.

\end{abstract}

%% file: intro.tex
\begin{figure*}
    \centering
    \includegraphics[width=\linewidth]{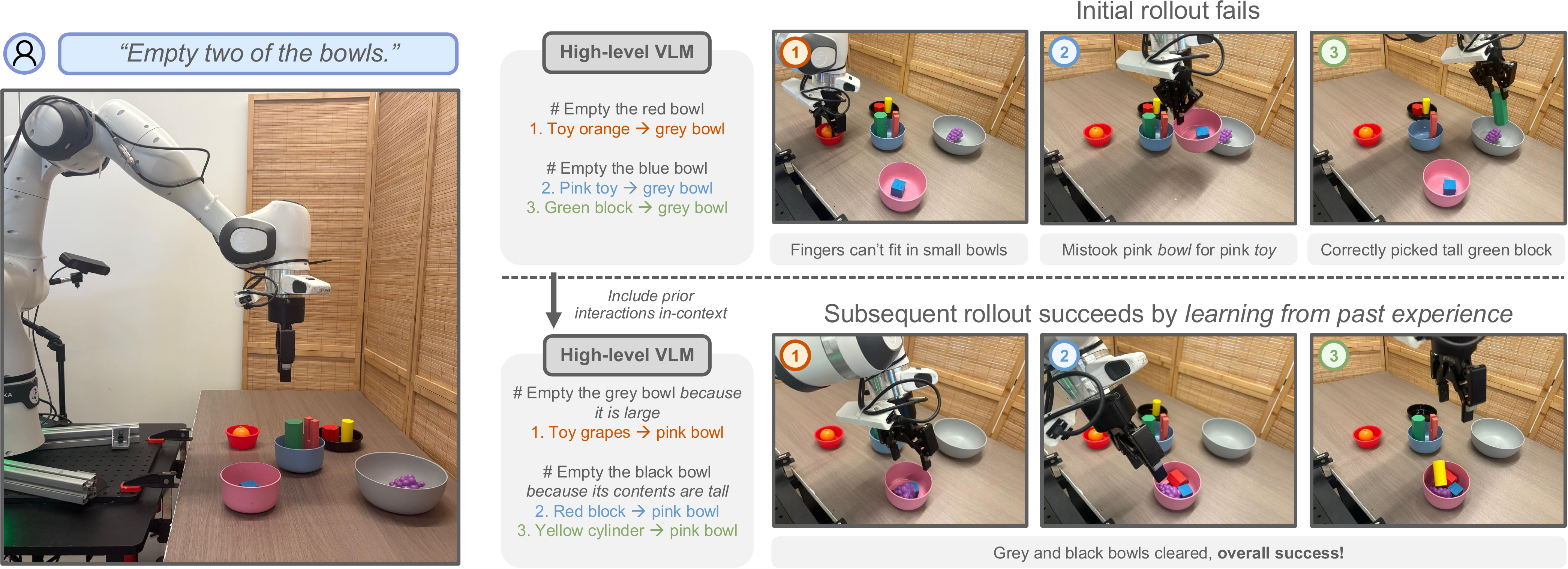}
    \caption{Our method, \textit{Learning from Inference Time Execution} (\method{}), is a novel approach that leverages a two-level hierarchical model to reason about complex tasks based on prior experience collected at inference-time, allowing it to learn the robot's affordances and gradually improve performance.}
    \label{fig:liten_teaser}
\end{figure*}

Robotic foundation models based on powerful pre-trained vision-language model (VLM) backbones have the potential to combine both the semantic and common-sense problem-solving abilities of LLMs and the flexible and dexterous end-to-end control capabilities of learned policies~\cite{ahn2022can, brohan2023rt, black2024pi_0, kim2024openvla, octo_2023}. However, current robotic foundation models, most notably \textit{Vision-Language-Action} models (VLAs), have primarily been studied in ``single shot'' settings, where they are evaluated on their ability to follow individual user commands. A practical robotic system needs to also plan through complex behaviors and, perhaps most importantly, adjust its behavior based on context and perceived capabilities. For example, if the robot needs to open a latched container, it might try to unlatch it in a particular way, and if that fails, it should modify its strategy and try a different approach. This kind of in-context adaptation has been observed as an \emph{emergent} behavior in LLMs~\cite{madaan2023self, gou2023critic, shinn2023reflexion}, but has proven difficult to enable in the robotics domain with current VLAs.


In this paper, we propose an inference-time learning method that enables robotic policies to reason through complex tasks, form plans, and then adjust those plans based on observed outcomes (without any additional training). Our method, \textbf{Learning from Inference-Time Execution (\method)}, employs an off-the-shelf VLM to first attempt a task by controlling a low-level VLA, then retrospectively reason about the induced robot behaviors. These insights can then be included in-context in future attempts, allowing the VLM to progressively learns \textit{affordances}~\cite{gibson2014ecological}: what the robot can or cannot do, subject to the constraints of dynamics, the robot embodiment, and the policy's learned behaviors. \method{}'s high-level VLM ``feels out'' the policy's capabilities, gradually strengthening its interface with the low-level controller and \textit{improving its high-level task reasoning as experience is accumulated}. We illustrate how \method\ can utilize past experiences in Figure~\ref{fig:liten_teaser}.


Our approach can be seen as an adaptation of inference-time self-refinement approaches for agentic foundation models~\cite{shinn2023reflexion, tong2024can, an2023learning} to real-world robotics. As shown in Figure~\ref{fig:liten_diagram}, \method\ adopts a two-phase iterative approach to self-refinement: a \textit{reasoning} phase, followed by an \textit{assessment} phase. In the initial reasoning phase, our high-level VLM has no prior experience of how the VLA will act when instructed, and must ``guess'' a plan (i.e., a sequence of subtask instructions). After the plan is attempted in the physical world, we move to the assessment phase and enlist a VLM \textit{judge} to evaluate the result of the VLA's execution. Unlike existing self-refinement methods that can leverage precise feedback in simulated or computational environments, we must deal with unstructured data (e.g., raw videos) and draw meaningful conclusions for use in refining robot task reasonings.




To gain valuable takeaways from unstructured videos of robot executions, \method\ breaks the process of assessment down into a sequence of steps that are individually manageable by the VLM judge. Specifically, the judge assesses the outcomes of robot rollouts by determining which subtask commands failed, what the policy did incorrectly, possible reasons why, and how these errors might be avoided. These insights are then provided in-context when the VLM next attempts the task, allowing it to revise its task reasoning in light of its prior experience.

In summary, our contribution is \method, a novel inference-time technique for high-level VLMs to solve complex robotic tasks by learning affordances through repeated interactions with the environment, which is leveraged for better task reasoning. Crucially, \method\ requires no additional training and can be used by off-the-shelf VLMs. We demonstrate the efficacy of \method\ on a set of long-horizon manipulation tasks using the DROID Franka robot setup~\cite{khazatasky2024droid}. Our results show that our approach is able to iteratively improve its planning capabilities from repeated attempts at a task, outperforming baseline approaches to inference-time learning that are not designed for real-world robotics.

%% file: related.tex

\textbf{Vision-Language-Action models.} Vision-Language Models (VLMs) are a powerful source of representations for control. They are popularly leveraged via behavioral cloning (BC), which yields Vision-Language-Action policies~(VLAs): VLMs that have been fine-tuned on vast robot demonstration corpora~\cite{openxembodiment2024, walke2023bridgev2, khazatasky2024droid, jiang2025dexmimicgen} into robot policies~\cite{brohan2023rt, kim2024openvla, black2024pi_0, bjorck2025gr00t}. VLAs are often generalists: due to their diverse training data, they can perform many tasks in unstructured settings, typically as commanded in text. Despite this, current VLAs are mainly trained on low-level tasks, and thus fail at high-level tasks that demand reasoning or common sense. While hierarchical~\cite{Pi2025pi05, shi2025hirobot} and reasoning~\cite{GRT2025geminiRobotics, zawalski2024ecot, chen2025ecot-lite, lin2025onetwovla} approaches alleviate this, such methods require expensive data collection, annotation, and training. In contrast, \method{} uses an off-the-shelf VLM to learn from inference-time rollouts by storing and reasoning about them in-context, without training.

\textbf{Planning and affordance reasoning.} 
An alternative approach that addresses this shortcoming is to use pre-trained foundation models as high-level planners or reasoners, typically interfacing with a separate low-level policy or controller~\cite{huang2022inner, zeng2022socratic}. These methods are \textit{zero-shot}: they forgo robot-specific training and instead use the common-sense knowledge and reasoning capabilities in VLMs to determine (1) what the robot \textit{should} do (e.g., by decomposing a goal into subtasks) and (2) what the robot \textit{can} do, grounded in the scene and the robot's capabilities (also called ``affordances''~\cite{gibson2014ecological}). This dichotomy is outlined by SayCan~\cite{ichter2022saycan}, which uses a language model to determine appropriate subtasks, while a separate learned value function estimates affordances. Other works learn a value function estimation for foundation-model based approaches either in an offline reinforcement learning setting~\cite{nakamoto2024steering} or through visual heuristics~\cite{ma2025valuefxn}.
In contrast, VoxPoser~\cite{huang2023voxposer} leverages (V)LMs to determine affordances by constructing a semantic voxel map in code. Other approaches couple affordance estimation and task reasoning more tightly, using the VLM both to interpret the scene and decide on appropriate behaviors~\cite{liang2023code, fangandliu2024moka}.

However, contrasting \method{}, the zero-shot nature of these approaches can also be a drawback: because the high-level VLM is not trained on robot data, it has limited understanding of the robot's capabilities and limitations, leading to suboptimal performance~\cite{ichter2022saycan}. This is only exacerbated as considered tasks become more complex, necessitating both more flexible low-level controllers and the VLM to have a more fine-grained grasp of possible behaviors.

\textbf{Inference-time learning for agentic tasks.}
Outside of robotics, iterative inference-time refinement methods have been explored for agentic tasks in other domains~\cite{shinn2023reflexion, tong2024can, an2023learning}. Such works often have similar considerations as in robotics, especially when applied to control in simulated environments~\cite{du2023guiding, wang2023voyager, bhat2024grounding}. However, translating these methods directly to real-world robot control can be difficult. Specifically, existing works (1) abstract away low-level control by delegating it to a powerful planner (often leveraging privileged simulation data), and (2) makes use of that same ground-truth simulator state for expressing agent observations, skills, and even interaction feedback in text. Both these assumptions cannot be made for real-world robotics. We compare \method\ with a direct adaptation of Reflexion~\cite{shinn2023reflexion} to illustrate our method's comparative efficacy in unstructured real-world robot manipulation applications.

%% file: problem.tex
We consider the regime of robotic control for long-horizon instruction following. We define a \textit{task} $\tasklanguage$ as a natural language description of the desired instruction, e.g., ``Empty two of the bowls'' from Figure~\ref{fig:liten_teaser}.

In order to accomplish free-form natural language instructions like this, we make use of language-conditioned policies $\pilow\left( a \mid o, \subtasklanguage \right)$, which map observations $o$ and input instruction text $\subtasklanguage$ to a distribution over low-level robot actions $a$. We specifically use \textit{vision-language-action} models (VLAs) as $\pilow$, as they are an effective end-to-end approach for learning generalist language-conditioned control policies \cite{brohan2023rt, black2024pi_0, kim2024openvla}.

Critically, while VLAs are open-vocabulary policies and can condition on arbitrary language, they often struggle when given open-ended, long-horizon instructions. Instead, they excel at the shorter-horizon, atomic behaviors found in their training data. To accomplish long-horizon tasks, they also need separate high-level policies to break the overall task into behaviors that the VLA is trained for~\cite{Pi2025pi05, shi2025hirobot}. 
To address this limitation, we use a highly-capable off-the-shelf VLM for high-level reasoning, $\pihigh$, decomposing the overall task $\tasklanguage$ into a sequence of subtask instructions $\subtasklanguage_0 \dots \subtasklanguage_n$. The VLA can then execute each subtask in sequence. When rolled out via a standard action-perception loop, this yields a subtask trajectory $\traj_i = \{(o_0, a_0)_i, (o_1, a_1)_i, \dots (o_T, a_T)_i\}$ (where actions are sampled from the VLA conditioned on the current observation and subtask language $a_t \sim \pilow(\cdot \mid o_t, \subtasklanguage_i)$ and $T$ is the length of the subtask trajectory). The overall trajectory is simply the concatenation of all these subtask trajectories $\traj_i$ in sequential order. The overall task $\tasklanguage$ is successful if it is accomplished by the end of the last subtask trajectory.

Finally, we note that the VLA has learned to map commands to a diverse range of low-level actions. While we have a sense of what language the VLA has been trained to follow (which is useful for giving examples of ``reasonable'' subtask language $\subtasklanguage_i$ for $\pihigh$ to output), we do not have any a priori description of its affordances (e.g., in what scenarios it is able to accomplish certain commands, and with what likelihood). We thus consider an \textit{iterative multi-round setting}, wherein the system can attempt the task multiple times. Critically, the system can pass forward some information to subsequent attempts. Our experiments thus compare different ways to use pre-trained VLMs as $\pihigh$, specifically showing that our approach, \method{}, is an effective way to infer affordances and improve performance from past robot rollouts.

%% file: methodology.tex
\begin{figure}
    \centering
    \includegraphics[width=\linewidth]{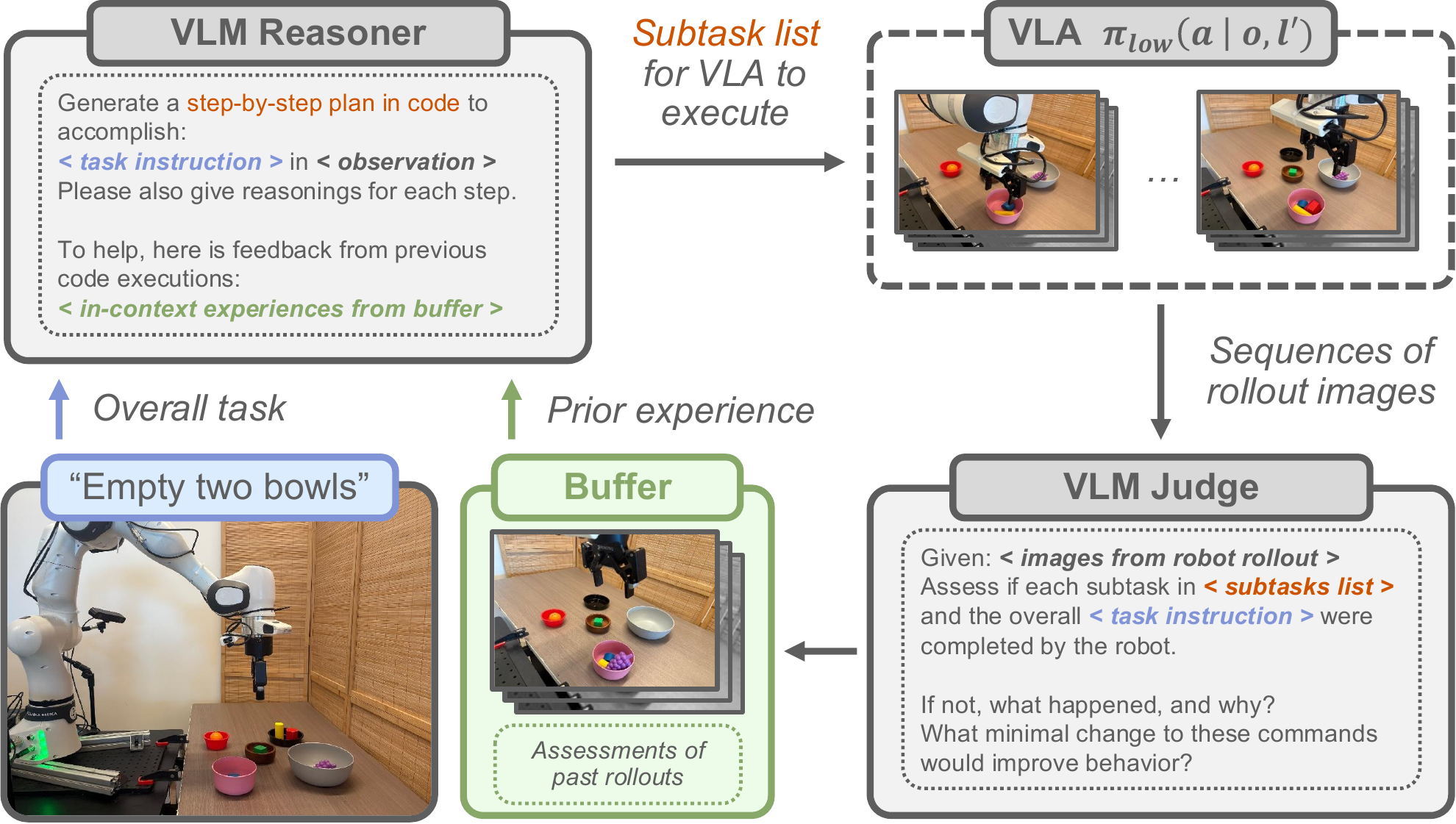}
    \caption{An overview of our approach. \method\ cycles between \textbf{(1) the reasoning phase}, wherein the VLM reasons about the task to decompose the task into subtasks that the VLA low-level policy can roll out in sequence and \textbf{(2) the assessment phase}, wherein the VLM judges the outcomes of inference-time rollouts. The reasoning phase includes the outputs of the assessment phase by including them in the VLM's context, allowing the reasoner to iteratively learn robot affordances and improve performance.}
    \label{fig:liten_diagram}
\end{figure}

We now present \method, an approach that allows the VLM planner $\pihigh$ to learn the affordances of the VLA $\pilow$ at inference time through repeated interactions with the physical world, subsequently using these affordances to improve its planning capabilities. We show that, by reasoning about prior attempts at a task -- how they succeed or fail, the underlying semantic or physical constraints that lead to these outcomes, and what sorts of behaviors the policy is empirically suited to -- \method{} is an effective way to gradually improve task success rates as more experiences are accumulated in-context, all \textit{with no extra policy training}.

\method{} consists of an iterative loop with two phases: a \textit{reasoning} phase and an \textit{assessment} phase. In the reasoning phase (Section~\ref{subsec:reasoning_phase}), the VLM \textit{reasoner} is given a task instruction $\tasklanguage$ and an initial image observation of the environment. It is instructed to reason through and decompose this goal into subtasks, which the VLA then executes as described in Section~\ref{sec:problem}. We then move to the assessment phase (Section~\ref{subsec:assessment_phase}), where we evaluate the outcome of each subtask execution in a structured format, invoking a VLM \textit{judge} with a chain of prompts to draw meaningful conclusions from unstructured trajectories. Finally, the result of this assessment is then added back in-context to the prompt for the VLM reasoner in the next iteration, which generates a new plan that takes the prior experience into account (Section~\ref{subsec:feedback_usage}). We provide a diagrammatic representation of \method\ in Figure~\ref{fig:liten_diagram}. We now detail each phase of \method.

\subsection{Reasoning Phase: Generating and executing plans} 
\label{subsec:reasoning_phase}

The reasoning phase begins by asking a VLM to act as $\pihigh$. We give it the overall task $\tasklanguage$ and initial observation image, then prompt it to produce a list of subtasks that solve this task, along with a justification for each step. 

To help direct $\pihigh$'s outputs, we supply the VLM reasoner with additional context. 
First, we give it provide a description of language commands that are representative of instructions in the VLA's pre-training data. This constrains $\pihigh$'s outputs to the ``style'' of language subtasks that $\pilow$ can follow (though critically \emph{not} its affordances, i.e., when each command is feasible).
We also describe the general class of instructions that are relevant to our task setup: in our setting, these are pick-and-place and move operations by a robot arm. Lastly, to ground the reasoner in its environment, we employ a similar technique to prior work~\cite{chen2022open, smith2025steer} and provide it with a list of manipulable objects, identified by a separate VLM call. Executing each subtask $\subtasklanguage_i$ in sequence with the VLA yields a corresponding trajectory segment $\traj_i = \{ (o_0, a_0)_i, (o_1, a_1)_i, ... \}$.

\subsection{Assessment Phase: Learning from previous executions}
\label{subsec:assessment_phase}
Once a plan has been fully executed, we begin the assessment phase by collecting the sequence of sub-trajectories for each subtask attempted by $\pilow$, i.e., $\{ (\ell'_0, \traj_0), \dots (\ell'_n, \traj_n)\}$. To generate useful feedback for future iterations, we present a \textit{structured assessment procedure} that asks a VLM judge to answer a chain of prompts about each subtask that mirrors human reasoning. The prompts are as follows:
\begin{enumerate}
    \item \textit{Did it succeed?} We give a subtask instruction $\ell'_i$ and the first and last observation images of $\traj_i$ as input to the judge, and ask whether or not the robot successfully accomplished $\ell'_i$. 
    \item \textit{What happened instead?} In the case of failure, we ask the judge to describe what the robot did in the environment instead of $\ell'_i$. In this step, we experimented with providing a video (sampled at various frame rates) of $\traj_i$ but found that our VLMs struggled to accurately comprehend unstructured sub-trajectory videos. For our experiments, we found it sufficient to provide the first and last observation images when asking the judge to describe what happened.
    \item \textit{Reason about failure.} We give the intended subtask $\ell'_i$, the description of what actually happened from the previous step, and the initial observation image to the judge and ask it to reason about why $\pilow$ failed in the way it did. To ensure that the judge can correctly diagnose failure causes, we include substantial context in our prompt that outlines a general VLA training and inference procedure and briefly describes the way $\pilow$ attends to language (e.g., `Our VLA tends to pay attention to the color, shape, and spatial orientation of objects in the instruction.') We additionally ask the VLM to suggest what \textit{minimal} changes to either $\subtasklanguage_i$ or the environment would improve the chance of success. We emphasize minimality to encourage suggested changes that are grounded in the specific task context. 
\end{enumerate}

\begin{figure*}
    
    \centering
    \includegraphics[width=\linewidth]{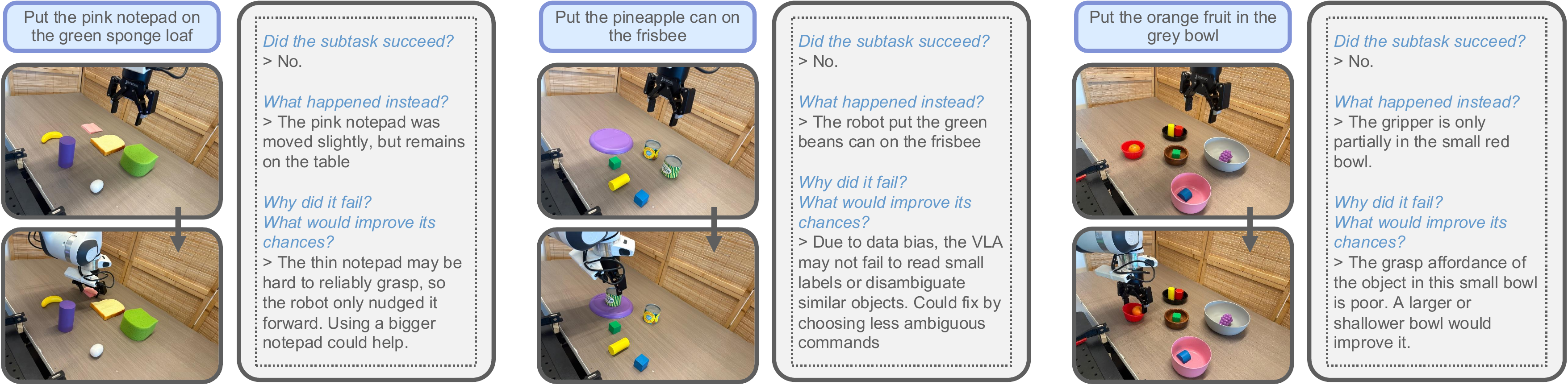}
    \caption{Examples of VLM judge's prompt and output for three subtasks (abridged for length). These generations are included in-context for the VLM reasoner in subsequent iterations attempting the task. Additional full examples are available in our codebase at https://github.com/ameesh-shah/liten-vla}
    \label{fig:liten_judge_example}
\end{figure*}

We provide an example illustrating the above steps in Figure~\ref{fig:liten_judge_example}.
In the case of a successful subtask, we instead ask a VLM to briefly describe the environment in which the success occurred. Once we have generated assessments for each subtask, we then ask a VLM to evaluate the overall task. We provide the structured assessments for all subtasks in an execution, along with overall task language $\tasklanguage$, and ask the judge to describe why the overall task succeeded or failed based on the outcomes of each subtask.

\subsection{Using feedback in future iterations} 
\label{subsec:feedback_usage}
Upon the start of the next reasoning phase, the structured feedback generated in all previous assessment phases is added in-context to the plan generation prompt. To ensure that assessments are used effectively, we add instructions on how to use prior experience to the VLM reasoner's prompt. In the prompt, we emphasize that a subtask's success is dependent on the environment, and that subtasks with multiple occurrences in-context do not necessarily share the same environment configuration with one another nor the current environment. Our prompt also notes that the VLA is stochastic, and its performance on the same subtask may vary across executions. We instruct the VLM reasoner to prioritize using subtask instructions in the following order: (1) instructions that are very likely to succeed in the current environment based on prior experience, (2) instructions that have not yet been tried, and (3) instructions that are unlikely to succeed but have been rephrased in a way that maximizes their success likelihood. The VLM reasoner takes into account its prior experience and the aforementioned instructions, then generates a new plan for execution. 

\subsection{Implementation}
\label{subsec:implementation}

We now describe our concrete instantiation of \method\ on a real-world robotics setup. We use GPT-5-mini~\cite{openai2025gpt5} as our high-level VLM in both the reasoning and assessment phases, given its cost effectiveness and state-of-the-art reasoning capabilities. For our low-level policy, we use $\pi_{0.5}$-DROID~\cite{Pi2025pi05}, an open-source state-of-the-art VLA that has been fine-tuned on the DROID  dataset~\cite{khazatasky2024droid}. Accordingly, our experiments use the standard DROID robot setup: a tabletop with a 7-DoF Franka Emika Panda robot arm and a 2F-85 Robotiq gripper end-effector operating at 5Hz.

The reasoning phase takes as input a user-specified $\tasklanguage$, and an initial image of the scene that comes from a fixed ZED 2.0 camera on the right side of the tabletop. The reasoning phase is a single VLM request that includes the following in addition to the aforementioned inputs: the top-level plan generation prompt for $\tasklanguage$, the in-context experience collected from prior attempts, and the instructions on how to use past experience.
The VLM generates Python code with comments explaining its reasoning to serve as the plan and uses a provided API endpoint that allows it to invoke the VLA for a subtask $\subtasklanguage_i$. Additionally, we ask the VLM reasoner to justify its subtask generations in natural language.

During execution, $\pi_{0.5}$-DROID receives joint angle proprioception and RGB images from the same right-side ZED camera and a ZED mini wrist camera. Subtasks are run with a maximum horizon of 300 time steps. $\pi_{0.5}$-DROID produces joint velocity action chunks~\cite{zhao2023chunking} of length 8, so the VLA only runs inference every 8 steps. The robot's position is reset to home between subtasks.

In the assessment phase, the initial and final image observations from each subtask are provided to the chain of prompts in our structured assessment template. Each step outlined in Section~\ref{subsec:assessment_phase} is a separate VLM request that uses the output from the previous steps. The results of all subtask assessments are then provided in a single request to the VLM judge when assessing the overall task. The overall task and subtask assessments, including the images used in those assessments, are then stored and provided in a hierarchical structure (subtasks are coupled with their overall task) in-context to future reasoning phases. Our implementation of \method, including the full prompts used, is available in our 
\href{https://github.com/ameesh-shah/liten-vla}{codebase}.

%% file: experiments.tex
We evaluate \method\ on a collection of challenging multi-step tasks that requires the high-level VLM to learn relevant affordances, including (1) what the low-level robot controller is capable of and (2) spatial and physical properties of the environment and embodiment. Through our experiments, we seek to answer the following questions:
\begin{enumerate}[leftmargin=1.3cm]
    \item[\textbf{RQ.1.}] Can \method\ learn task-relevant affordances in its environment by interacting with the physical world?
    \item[\textbf{RQ.2.}] As it gains experience, does \method\ effectively learn from both successes and failures, and iteratively improve its plans for complex tasks?
    \item[\textbf{RQ.3.}] What is the importance of each step in the assessment phase of \method\ in facilitating learning from past experiences?
\end{enumerate}

\subsection{Experimental Setup}
\begin{figure*}[t]
\centering
    \includegraphics[width=\textwidth]{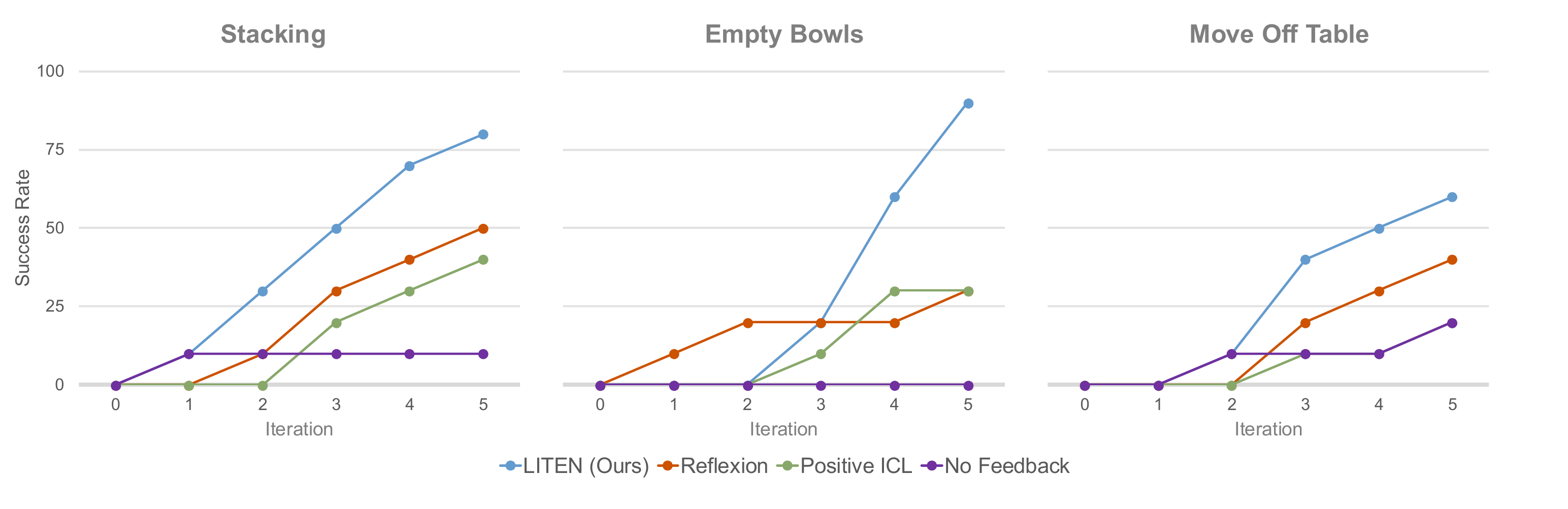}

    \vskip -0.5em
    \centering
    \caption{Success rates for the full completion of each multi-stage task over five attempt iterations. Our results show that \method\ is able to make effective use of prior attempts and consistently improves its plans as more experience is collected. Results are averaged over 10 trials.}
\label{fig:success_plots}
\end{figure*}



\begin{figure}
\centering
        \par
        \includegraphics[width=0.9\linewidth]{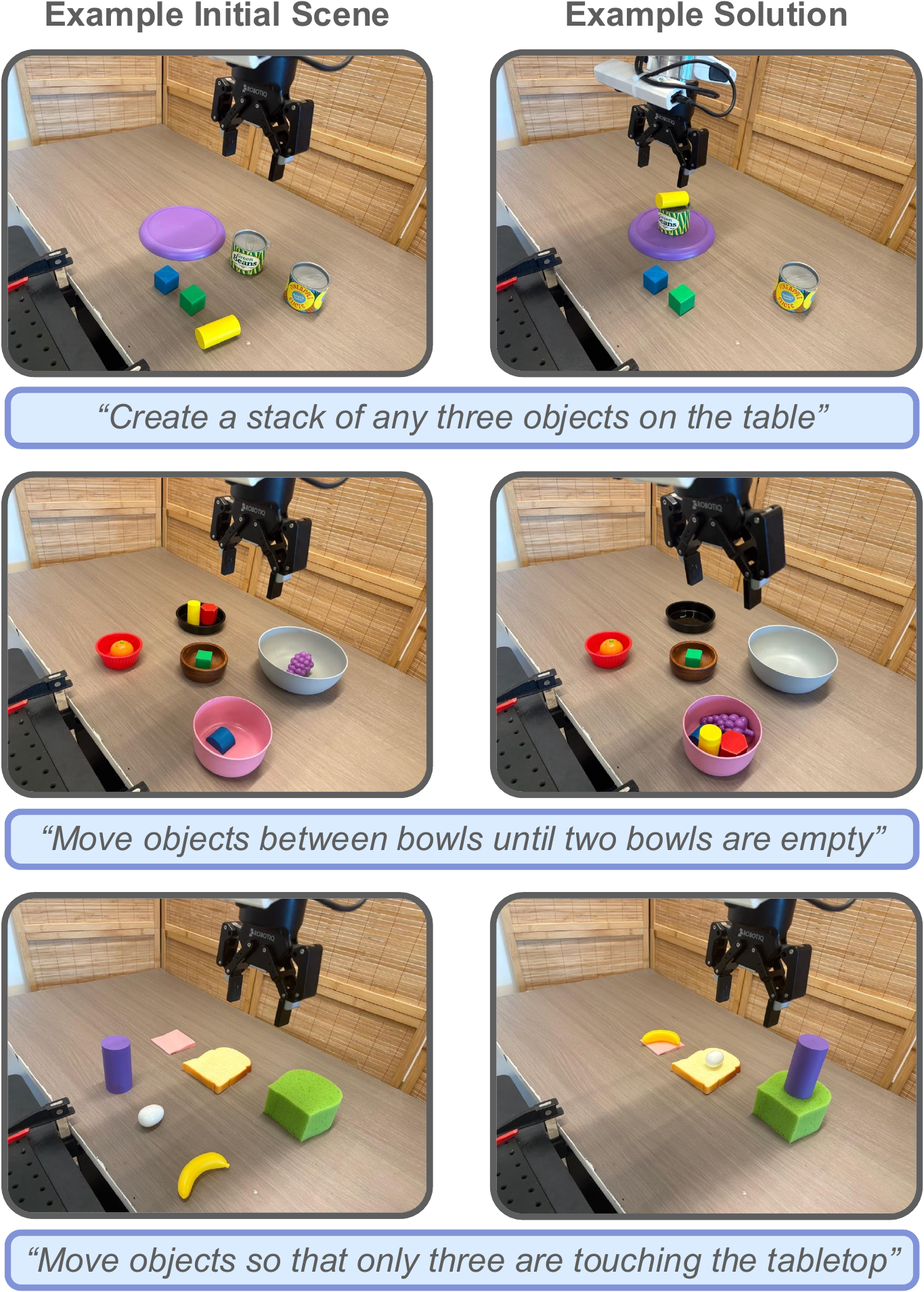}
    \vskip -0.5em
    \caption{Examples of initial configurations and solutions for our tasks.}
\label{fig:task_examples}
\end{figure}

We select three complex, multi-stage tasks that are achievable by composing behaviors and instructions akin to those found in the DROID dataset. To ensure that our VLA can consistently achieve our tasks, we fine-tune $\pi_{0.5}$-DROID on a small number of collected demonstrations to help it adapt to our specific tabletop and task setups. We collect 150 total demonstrations per multi-stage task, which are divided up into 5-15 distinct subtasks. Additional details are provided in section~\ref{subsec:additional_impl}. The initial configuration of our task does not vary across iterations in a given trial, but does vary slightly across trials; e.g., some object initial positions are switched.
Our tasks are depicted in Figure~\ref{fig:task_examples} and described as follows:
\begin{enumerate}
    \item \textbf{Stacking.} The robot must stack a total of three objects atop one another. The scene has 3 small blocks, 2 medium-sized cans, and a large upside down plate. The high-level VLM must learn which objects can be easily stacked atop one another by the VLA and which objects are too difficult to precisely balance.
    \item \textbf{Emptying Bowls.} A number of bowls of different depths and sizes are set on the table. The robot must empty two bowls by moving their contents to other bowls. The high-level VLM must learn which bowls are either large enough for the gripper end-effector to fit in \textit{or} shallow enough to allow objects to protrude and be grasped by the arm.
    \item \textbf{Moving Off Table.} The robot must move objects on the table onto other objects such that only three objects remain in direct contact with the table. The high-level VLM must learn which objects are manipulable by the VLA, and which objects can serve as landing spots for other objects without rolling or falling off.
\end{enumerate}

\textbf{Baselines.} We compare our approach against a number of alternative approaches to learning from inference-time experience that do not reason about experiences in the same way as \method.
While to our knowledge prior VLA methods do not utilize inference-time experience in-context, we adapt \textit{Reflexion}~\cite{shinn2023reflexion}, an inference-time learning technique for agentic LLMs, to a robotic setting. Our adaptation `reflects' on unstructured videos of executed plans, and uses these reflections in-context for subsequent iterations. This baseline can be thought of as a na\"{i}ve adaptation of standard self-refinement approaches to real-world robotics. We also compare against an approach that only stores successful subtask attempts and provides those successful attempts in-context for future plan generations, called \textit{Positive-ICL}. This approach mirrors existing methods that use positive in-context examples for zero-shot open-world robotic manipulation~\cite{liang2023code}, and can be viewed as an ablation of our method that does not use negative examples. Lastly, we compare against a na\"{i}ve baseline that regenerates a new plan at every iteration without using any feedback, called \textit{No-Feedback}. 

\textbf{Ablations.} We answer \textbf{RQ.3.} by conducting an ablation study where we remove steps from the assessment phase (see Section~\ref{subsec:assessment_phase}) and run \method\ in the same manner as described above for the Empty Bowls task. We ablate the approach in two ways: (1) removing the failure reasoning (\textbf{w/o failure reasoning}: ``why did the policy fail at this subtask?'') and (2) removing the outcome analysis as well (\textbf{w/o failure reasoning and outcome}: ``what did the policy do instead?''), leaving only whether the subtasks were successful or not.

\subsection{Results}

\textbf{Main results.} In Figure~\ref{fig:success_plots}, we show the results of each approach in accomplishing the tasks over a total of five iterations, averaged over ten trials. For each trials, we run each method for a maximum of five iterations, stopping early if the attempt is successful.
The success rates represent the rate of accomplishing the entire task and do not consider partial credit.
Our results show that \method\ is able to make use of its collected experience and successfully instruct $\pilow$ with high-affordance subtask instructions to accomplish the overall tasks, answering \textbf{RQ.1.} and \textbf{RQ.2.} affirmatively. \method's success rate increases consistently over consecutive attempts, indicating that it effectively uses additional prior experience to further improve its plans. The baseline approaches are substantially less effective at leveraging prior experience. Notably, the na\"{i}ve No-Feedback baseline is practically unable to accomplish any tasks, showing that the VLM will not occasionally generate a correct plan as a result of random chance. This shows the importance of learning from past experience at inference time.

\begin{figure}
\centering
        \par
        \includegraphics[width=0.98\linewidth]{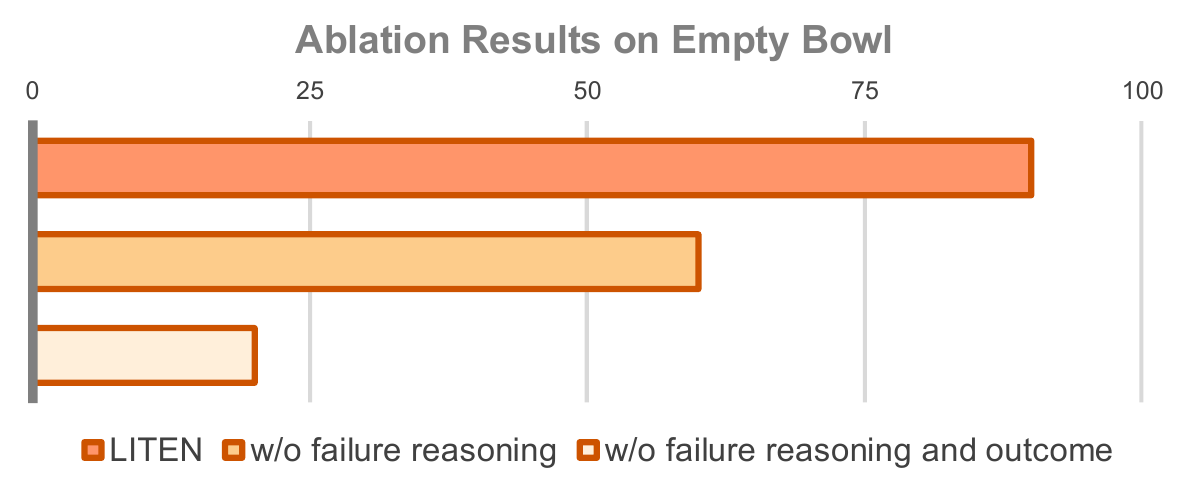}
    \vskip -0.5em
    \caption{Success rates after five past attempts after ablating various components of \method{}'s VLM judge. When any part of the assessment phase is ablated, performance drops dramatically, indicating that prior experience is most impactful when it includes whether the past rollout succeeded or not, what the policy did instead if not, and why that might be the case.}
\label{fig:ablation_bars}
\end{figure}

\textbf{Ablation results.} We present the performance of \method{} under various ablation conditions in Figure~\ref{fig:ablation_bars}, demonstrating the importance of the knowledge gained in each step of the assessment phase. Providing only the success or failure of each subtask (without failure reasoning or outcome analysis) yields the worst performance. Removing the failure reasoning only performs better, but will on occasion attribute differences between intended and actual outcomes to the instruction's language as opposed to physical explanations. The explicit ``reason about failure'' step provides useful context about why failures may have occurred, such as suggesting the limitations of VLA capabilities or physical properties of objects, and thus allows \method{} to learn more sophisticated affordances.

\textbf{Qualitative analysis.}
We provide a qualitative analysis to answer \textbf{RQ.1.} and describe the affordances \method\ learns in our experiments.
We find that \method\ is particularly successful at learning from failures that either indicate (1) biases from the VLA or (2) physical properties that cause obvious difficulties in control. 

To exemplify (1), in the Stacking task, the VLA is biased towards manipulating larger objects in the scene, such as the two cans or the yellow cylinder block. In an attempt from our experiments, the high-level VLM instructed the VLA to ``put the green cube on the purple plate'', but the VLA instead moved the green beans can onto the purple plate. \method\ was able to clearly identify this mistake made by the VLA and recognize that moving the green can to the purple plate is a subtask that can be part of a potential solution. The subsequent plan included ``put the green beans can on to the purple plate''. 

Regarding (2), we found that the high-level VLM often suggested initial plans that require precise control. For the Stacking task, the high-level VLM would frequently generate a plan involving stacking the three small blocks instead of the larger cans and plate, which is comparatively more difficult and requires finer manipulation skills. In the Moving Off Table task, the VLM occasionally suggested moving the green sponge on top of other objects, despite it being the largest object on the table and too wide for the gripper. However, \method\ enabled the high-level VLM to draw meaningful conclusions from these attempts (e.g., ``the VLA may lack precise top-down placement abilities when stacking the blue and green blocks'' or ``the gripper could not grasp the green sponge due to size constraints'') and generate plans that adjusted accordingly.

The baselines are unable to learn as effectively. Because the Positive-ICL baseline only retains successful subtasks in-context, the method must effectively ``get lucky'' and find successful subtasks through the randomness of the VLM-generated plans, leading to high inefficiency. This illustrates how \method{}'s fine-grained rollout assessments critically provide a strong signal for understanding the robot's capabilities and leveraging it for improvement.
The Reflexion baseline reasons over entire raw video trajectories, as opposed to our structured reasoning approach, which we found often led to poor comprehension of physical outcomes. For example, we found that entire-video reflections would often hallucinate objects that were not in the scene, or mistakenly assume subtasks were successfully completed. This underscores the challenge of extracting meaningful feedback from raw robot interaction data.

\section{Discussion} 

\subsection{Failure Cases}

We characterize the failure cases of \method\ when it fails to correctly plan for a task after multiple execution attempts. The majority of failures can be attributed to either (a) the stochasticity of the VLA, (b) \method's tendency to attribute control failures to the wording of a subtask, or (c) \method's inability to reason causally about subtask orderings. 

For the first failure case, a small number of potential subtasks in our experiments have high variance, such as stacking one can atop another in the Stacking task. If \method\ experiences a successful subtask execution in an early iteration, it will tend to fixate on using this subtask, even if multiple subsequent executions of the same subtask fail. 

In the second case, if a subtask execution fails due to a control error, \method\ may misdiagnose this failure as a result of imprecise language in the subtask instruction. For example, if the subtask ``put the blue block on the green block'' fails, but the blue block gets moved close to the green block, \method\ may attribute this failure to the language and provide an instruction like ``put the blue block directly on top of the center of the green block.''

The third case is most easily understood via example. In the Move Off Table task, overall success is highly sensitive to the ordering and control precision of each subtask. In one instance, the purple cylinder was successfully placed on the green sponge. However, the next subtask was placed the white egg also on the green sponge, which, while successful, knocked the purple cylinder back onto the table in the process. Although the VLM judge managed to accurately describe this phenomenon during the assessment phase (``The VLA likely ... displaced the cylinder when placing the egg''), the VLM reasoner struggled to make use of this insight in future plans.

\subsection{Analysis and Limitations}
We now take a more prescriptive view of the cases discussed earlier.
Both the first and second failure cases represent failure of the VLM to adapt to VLA affordances on \textit{individual} subtasks. 
In the first case, the VLM reasoner needs to strike a balance between avoiding ``borderline affordances'' of VLAs, while attempting to leverage the VLA's steerability through language at the same time. 

The second failure case is an artifact of using language as the reasoning mode. We expect that as VLM video comprehension capabilities improve, the VLM judge can properly classify infeasible actions that no amount of subtask language modification will fix, mitigating this failure mode.

The last case represents what we see as the most salient current limitation of \method. 
Unlike the first two failure cases, it exhibits a sequence of successful individual subtasks but a failure at the overall task level. 
This indicates a deficiency in the ability of the VLM to reason across subtasks.
Thus, instilling the high-level VLM reasoner with stronger logical and causal reasoning is a clear area of future improvement for inference-time affordance learning.

%% file: conclusion.tex
In this work, we present \method, a novel inference-time learning method that enables generalist robot policies to reason about complex tasks, attempt them in the real world, and then distill insights from those past attempts to improve future reasoning, without requiring additional training. \method\ iterates through a two-phase process that first generates plans for a VLA using a high-level VLM, then assesses the execution of that plan through a chain of prompts that enables the VLM to draw meaningful conclusions for use in subsequent planning iterations. Our experimental results show that \method\ effectively uses past experience to solve complex manipulation tasks that require an understanding of the robot's affordances.

\method{} joins an emerging body of work on in-context learning for robotics~\cite{vosylius2024instant, fu2024context}. We see the broad applicability of \method{} as one of its key strengths: in addition to not requiring training, \method{} can be used with any off-the-shelf VLM and VLA. Further, our approach is hardware-agnostic, only requiring changes to our prompts for adaptation to new robot setups. In a similar vein, we expect \method{} to become increasingly useful as the capabilities of (robotic) foundation models improve: if VLMs can better understand the physical world and VLAs can more accurately attend to language instructions, \method{} will empower the combination of these stronger models to solve highly complex tasks.

There are many exciting directions for future work. Although \method\ is only demonstrated on single tasks, its usage in a \textit{lifelong learning} setting is compelling, where past experience across many different tasks is used to solve new tasks. To that end, we are interested in understanding what past experiences are more or less useful in efficiently gaining affordances.
We are also interested in scaling up \method\ to settings with large amounts of prior data that cannot entirely be fit in-context, which will require adapting context management techniques such as RAG~\cite{lewis2020retrieval}.

%% file: appendix.tex

\subsection{Additional Implementation Details}
\label{subsec:additional_impl}
The implementation of \method{} used in our experiments uses a fine-tuned version of the $\pi_{0.5}$~\cite{Pi2025pi05} VLA as our choice of low-level controller. $\pi_{0.5}$ demonstrates state-of-the-art generalization in natural language instruction following, due in large part to the vast set of heterogeneous data sources and tasks in its pre-training data. This data mixture includes demonstrations that involve step-by-step sub-task following, similar our usage of $\pi_{0.5}$ in our own experiments. 

To improve $\pi_{0.5}$'s ability to act in our specific experimental setup, we use $\pi_{0.5}$-DROID, a version of $\pi_{0.5}$ that has been fine-tuned on the full DROID dataset~\cite{khazatasky2024droid}, which contains demonstrations on the same hardware setup as ours. However, we found that using $\pi_{0.5}$-DROID without any additional training in our setup led to prohibitively poor performance. Specifically, we observed that $\pi_{0.5}$-DROID struggled to execute fine-grained operations required for pick-and-place instructions in our setting, such as properly orienting the gripper with the correct depth and angle to grasp small objects while avoiding others. 

We further improved $\pi_{0.5}$-DROID's capabilities so that it could feasibly achieve the long-horizon tasks in our experiments by fine-tuning $\pi_{0.5}$-DROID on a small set of sub-task demonstrations collected in our tabletop setting. These demonstrations were collected through teleoperation in a manner identical to how demonstrations were collected in the original DROID dataset~\cite{khazatasky2024droid}. For each experimental task, we collected 150 demonstrations. We did not collect demonstrations on task instructions that were either physically infeasible or extremely difficult, e.g., balancing a rectangular block on a cylindrical block. We also varied the initial layout of the tabletop for each collected demonstration to prevent overfitting (note that initial layouts from our demonstration set often differed significantly from the layout used during evaluation.)

We provide the specific task instructions collected along with the number of demonstrations collected for each instruction below:

\begin{promptbox}
\% Stacking \\
Put the green block on top of the blue block: 15 \\
Put the blue block on top of the green block: 15 \\
Put the green beans can on top of the purple plate: 20 \\
Put the green beans can on top of the yellow pineapple slices can: 15 \\
Put the yellow pineapple slices can on top of the green beans can: 15 \\
Put the yellow pineapple slices can on top of the purple plate: 15 \\
Put the yellow cylinder block on top of the green beans can: 20 \\
Put the yellow cylinder block on top of the yellow pineapple slices can: 20 \\
Put the yellow cylinder block on top of the purple plate: 15
\end{promptbox}

\begin{promptbox}
\% Empty Bowls \\
Move the green block to the pink bowl: 15 \\
Move the green block to the black bowl: 15 \\
Move the orange fruit to the pink bowl: 15 \\
Move the orange fruit to the gray bowl: 15 \\
Move the yellow cylinder block to the gray bowl: 15 \\ 
Move the yellow cylinder block to the pink bowl: 15 \\
Move the red hexagonal block to the gray bowl: 15 \\
Move the red hexagonal block to the pink bowl: 15 \\
Move the purple grapes to the pink bowl: 15 \\
Move the purple grapes to the gray bowl: 15
\end{promptbox}

\begin{promptbox}
\% Move Off Table \\
Put the purple cylinder on the green sponge: 30 \\
Put the purple cylinder on the foam bread slice: 15 \\
Put the white egg on the green sponge: 15 \\
Put the white egg on the foam bread slice: 30 \\
Put the white egg on the pink notepad: 10 \\
Put the yellow banana on the foam bread slice: 15 \\
Put the yellow banana on the pink notepad: 30 \\
Put the pink notepad on the foam bread slice: 5 \\
\end{promptbox}

We fine-tuned $\pi_{0.5}$-DROID on each collected demonstration set separately for each experiment. We fine-tuned for 2500 training steps, using a batch size of 128 and an action chunk size of 16. The learning rate used began at 3e-5 and decayed on a cosine schedule every 250 steps with a decay rate of 2e-6. We performed training on a set of four NVIDIA A100 GPUs.

\subsection{Additional Qualitative Analysis}

\begin{figure*}
    
    \centering
    \includegraphics[width=\linewidth]{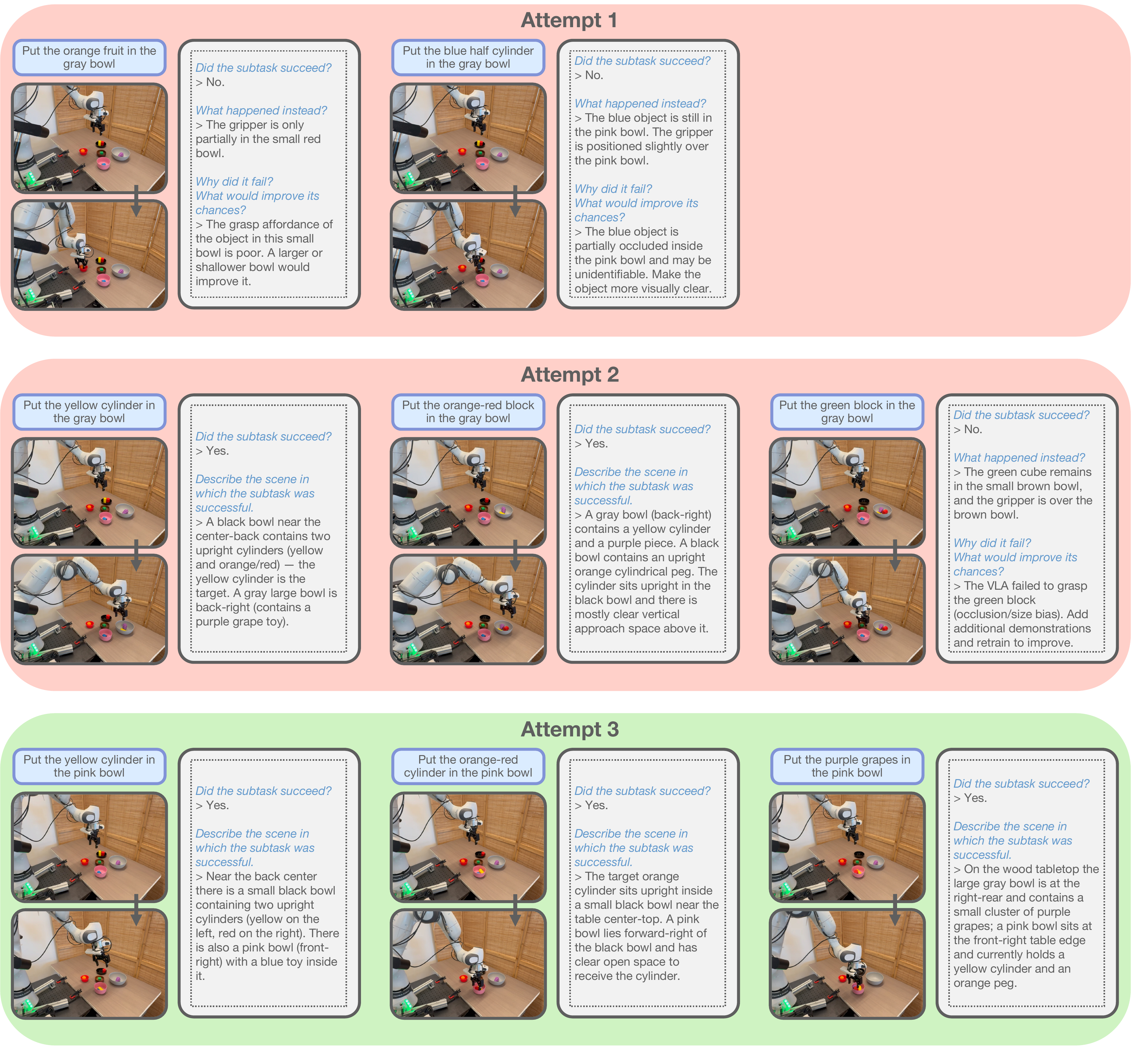}
    \caption{An example sequence of attempts for \method{} on the Empty Bowls task that resulted in a successful execution. VLM outputs from the reasoner and judge are condensed for length and readability.}
    \label{fig:liten_bowls_example}
\end{figure*}

\begin{figure*}
    
    \centering
    \includegraphics[width=0.8\linewidth]{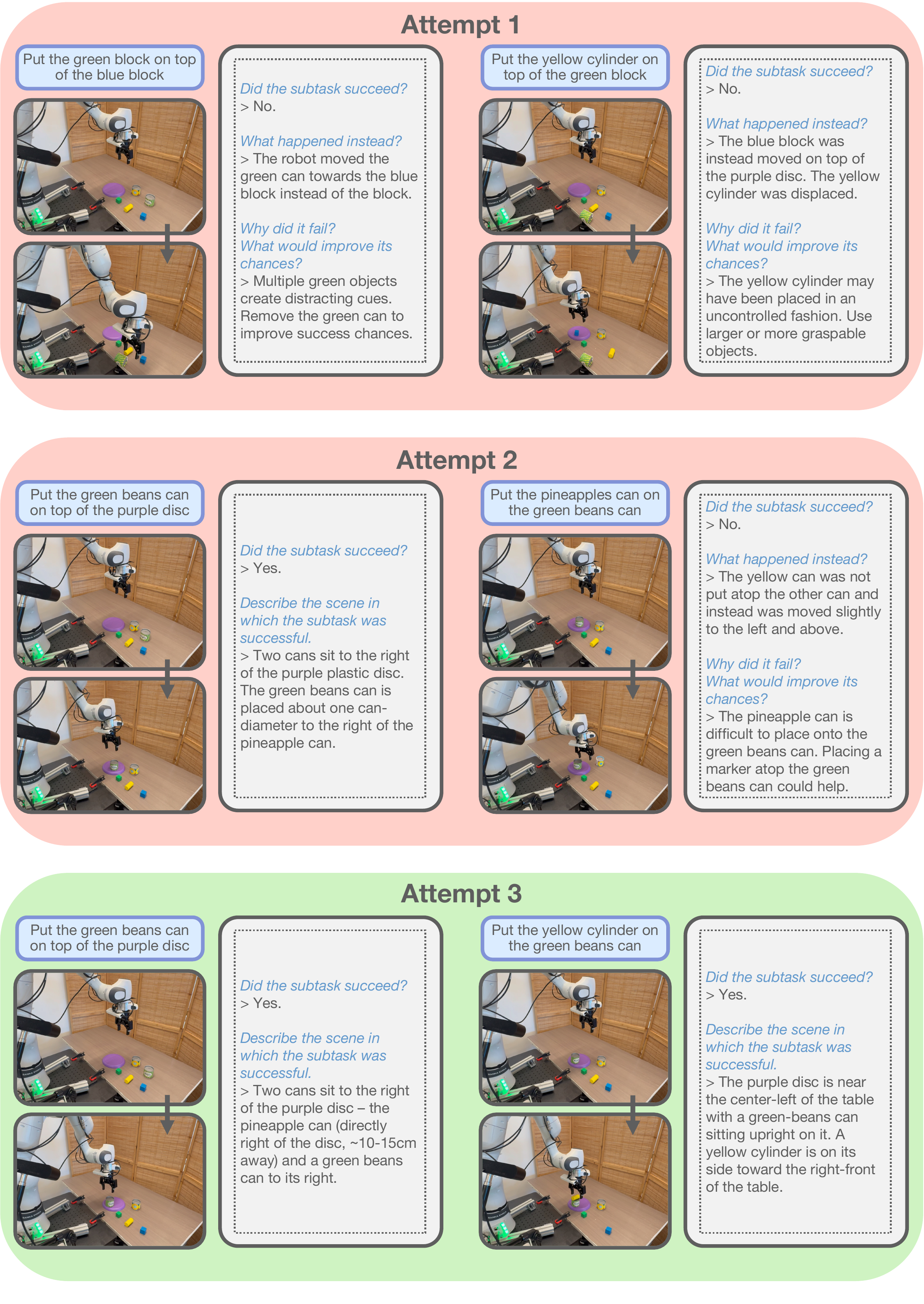}
    \caption{An example sequence of attempts for \method{} on the Stacking task from our experiments that resulted in a successful execution. VLM outputs from the reasoner and judge are condensed for length and readability.}
    \label{fig:liten_stack_example}
\end{figure*}

\begin{figure*}
    
    \centering
    \includegraphics[width=\linewidth]{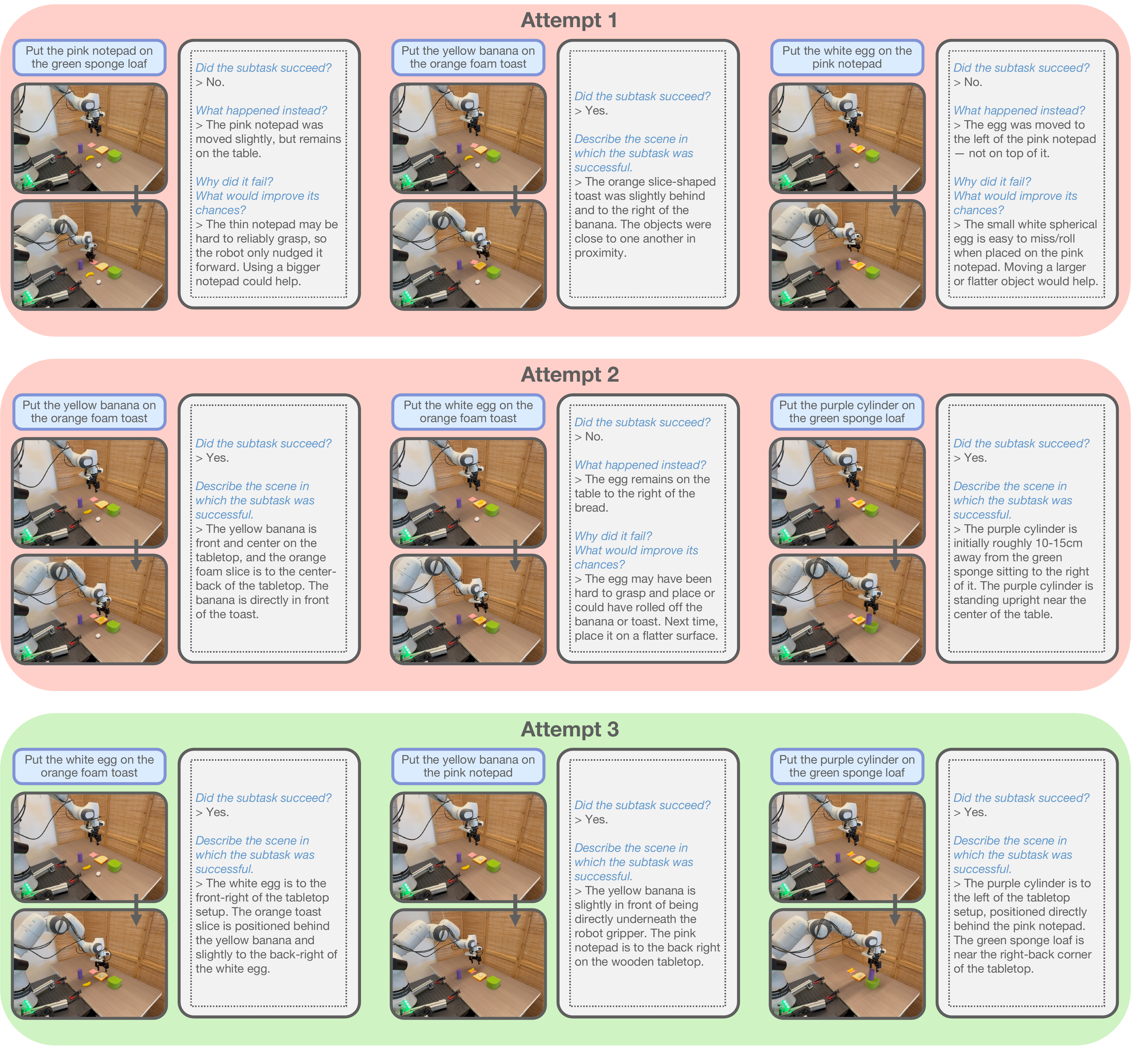}
    \caption{An example sequence of attempts for \method{} on the Empty Bowls task from our experiments that resulted in a successful execution. VLM outputs from the reasoner and judge are condensed for length and readability.}
    \label{fig:liten_offtable_example}
\end{figure*}

In this section, we provide examples of multiple attempt iterations generated by \method{} on each of our experimental tasks. The examples, provided in Figures~\ref{fig:liten_bowls_example},~\ref{fig:liten_stack_example} and~\ref{fig:liten_offtable_example}, illustrate the evolution of sub-task sequences generated by the VLM reasoner, along with feedback from the VLM judge produced during the assessment phase. We provide the initial and final images of the environment for each sub-task's execution that were given as input to the VLM judge during the assessment phase. For conciseness, we abridge the output from the VLM judge and include the most relevant text in our illustrations. The overall task-level reasoning performed after each attempt was also elided for brevity. Each of the provided examples ultimately culminates in a successful attempt by the VLM reasoner.

We identify additional observations from our evaluation of \method{}, highlighted by the provided examples. When asked to reason about the failure of a given sub-task, the VLM judge generates multiple possible hypotheses, which typically span across (1) potential inaccuracies in the sub-task's language instruction, (2) potential visual distractors or difficulties, (3) control errors either attributed to the VLA's capabilities or the dynamics of the environment, and (4) a lack of relevant data in the VLA's training or fine-tuning datasets. In our experiments, categories (2) and (3) are often the most valuable for the reasoner when generating the next sub-task sequence. For example, in the Stacking task example from Figure~\ref{fig:liten_stack_example}, the VLA fails to ``put the pineapples can on the green beans can'' in Attempt 2. Although the VLM judge suggests placing a marker on top of the green beans can to improve the chance of success, which is infeasible in our experimental setup, the VLM reasoner is able to utilize other parts of the judge's feedback, such as the difficulty of stacking two cans, and instead opts to stack the smaller, easier to place yellow cylinder in the third attempt.

Category (1), where the VLM judge suggests modifications to the sub-task's language instruction, were generally not useful in our experiments due to the language following capabilities of the fine-tuned $\pi_{0.5}$ VLA. Prior work has shown that current VLAs struggle to attend to fine-grained semantics in an instruction's language~\cite{glossop2025cast}, and we observed this to be the case in our evaluations as well. As VLAs continue to improve at language following, category (1) will become increasingly valuable for inference-time adaptation. In a similar vein, category (4) did not provide valuable feedback in the context of our experiments, given that our work studies a purely in-context learning setting.

As discussed in section~\ref{subsec:assessment_phase}, \method{}'s VLM judge assesses individual sub-tasks based on the first and last observation images of an execution, rather than (raw or downsampled) videos which led to poor performance. However, obvious limitations remain when using static images to assess a dynamic execution. Following the same example from the Stacking task in Figure~\ref{fig:liten_stack_example}, the second sub-task in Attempt 2 of stacking the pineapples can on the green beans can resulted in the robot placing the pineapple can partially on top of the green beans can, which led to the pineapple can falling back on to the table. The VLM judge, only able to see the before and after images, could only draw conclusions from seeing the pineapple can displaced from its original position, which led to less precise feedback than what otherwise would have been provided from an accurate interpretation of the video. As VLMs improve their ability to comprehend cause and effects in unstructured videos, \method{} will improve correspondingly.

%% file: root.bbl
\begin{thebibliography}{10}
\providecommand{\url}[1]{#1}
\csname url@rmstyle\endcsname
\providecommand{\newblock}{\relax}
\providecommand{\bibinfo}[2]{#2}
\providecommand\BIBentrySTDinterwordspacing{\spaceskip=0pt\relax}
\providecommand\BIBentryALTinterwordstretchfactor{4}
\providecommand\BIBentryALTinterwordspacing{\spaceskip=\fontdimen2\font plus
\BIBentryALTinterwordstretchfactor\fontdimen3\font minus \fontdimen4\font\relax}
\providecommand\BIBforeignlanguage[2]{{%
\expandafter\ifx\csname l@#1\endcsname\relax
\typeout{** WARNING: IEEEtran.bst: No hyphenation pattern has been}%
\typeout{** loaded for the language `#1'. Using the pattern for}%
\typeout{** the default language instead.}%
\else
\language=\csname l@#1\endcsname
\fi
#2}}

\bibitem{ahn2022can}
M.~Ahn, A.~Brohan, N.~Brown, Y.~Chebotar, O.~Cortes, B.~David, C.~Finn, C.~Fu, K.~Gopalakrishnan, K.~Hausman, \emph{et~al.}, ``Do as i can, not as i say: Grounding language in robotic affordances,'' \emph{arXiv preprint arXiv:2204.01691}, 2022.

\bibitem{brohan2023rt}
A.~Brohan, N.~Brown, J.~Carbajal, Y.~Chebotar, X.~Chen, K.~Choromanski, T.~Ding, D.~Driess, A.~Dubey, C.~Finn, \emph{et~al.}, ``Rt-2: Vision-language-action models transfer web knowledge to robotic control,'' \emph{arXiv preprint arXiv:2307.15818}, 2023.

\bibitem{black2024pi_0}
K.~Black, N.~Brown, D.~Driess, A.~Esmail, M.~Equi, C.~Finn, N.~Fusai, L.~Groom, K.~Hausman, B.~Ichter, \emph{et~al.}, ``$\pi_0 $: A vision-language-action flow model for general robot control,'' \emph{arXiv preprint arXiv:2410.24164}, 2024.

\bibitem{kim2024openvla}
\BIBentryALTinterwordspacing
M.~J. Kim, K.~Pertsch, S.~Karamcheti, T.~Xiao, A.~Balakrishna, S.~Nair, R.~Rafailov, E.~P. Foster, P.~R. Sanketi, Q.~Vuong, T.~Kollar, B.~Burchfiel, R.~Tedrake, D.~Sadigh, S.~Levine, P.~Liang, and C.~Finn, ``Openvla: An open-source vision-language-action model,'' in \emph{Conference on Robot Learning, 6-9 November 2024, Munich, Germany}, ser. Proceedings of Machine Learning Research, P.~Agrawal, O.~Kroemer, and W.~Burgard, Eds., vol. 270.\hskip 1em plus 0.5em minus 0.4em\relax {PMLR}, 2024, pp. 2679--2713. [Online]. Available: \url{https://proceedings.mlr.press/v270/kim25c.html}
\BIBentrySTDinterwordspacing

\bibitem{octo_2023}
{Octo Model Team}, D.~Ghosh, H.~Walke, K.~Pertsch, K.~Black, O.~Mees, S.~Dasari, J.~Hejna, C.~Xu, J.~Luo, T.~Kreiman, Y.~Tan, D.~Sadigh, C.~Finn, and S.~Levine, ``Octo: An open-source generalist robot policy,'' \url{https://octo-models.github.io}, 2023.

\bibitem{madaan2023self}
A.~Madaan, N.~Tandon, P.~Gupta, S.~Hallinan, L.~Gao, S.~Wiegreffe, U.~Alon, N.~Dziri, S.~Prabhumoye, Y.~Yang, \emph{et~al.}, ``Self-refine: Iterative refinement with self-feedback,'' \emph{Advances in Neural Information Processing Systems}, vol.~36, pp. 46\,534--46\,594, 2023.

\bibitem{gou2023critic}
Z.~Gou, Z.~Shao, Y.~Gong, Y.~Shen, Y.~Yang, N.~Duan, and W.~Chen, ``Critic: Large language models can self-correct with tool-interactive critiquing,'' \emph{arXiv preprint arXiv:2305.11738}, 2023.

\bibitem{shinn2023reflexion}
N.~Shinn, F.~Cassano, A.~Gopinath, K.~Narasimhan, and S.~Yao, ``{Reflexion: Language agents with verbal reinforcement learning},'' \emph{Advances in Neural Information Processing Systems}, vol.~36, pp. 8634--8652, 2023.

\bibitem{gibson2014ecological}
J.~J. Gibson, \emph{The Ecological Approach to Visual Perception: Classic Edition}.\hskip 1em plus 0.5em minus 0.4em\relax Psychology Press, 2014.

\bibitem{tong2024can}
Y.~Tong, D.~Li, S.~Wang, Y.~Wang, F.~Teng, and J.~Shang, ``Can {LLMs} learn from previous mistakes? investigating {LLMs}' errors to boost for reasoning,'' \emph{arXiv preprint arXiv:2403.20046}, 2024.

\bibitem{an2023learning}
S.~An, Z.~Ma, Z.~Lin, N.~Zheng, J.-G. Lou, and W.~Chen, ``Learning from mistakes makes {LLM} better reasoner,'' \emph{arXiv preprint arXiv:2310.20689}, 2023.

\bibitem{khazatasky2024droid}
\BIBentryALTinterwordspacing
A.~Khazatsky, K.~Pertsch, \emph{et~al.}, ``{DROID:} {A} large-scale in-the-wild robot manipulation dataset,'' in \emph{Robotics: Science and Systems XX, Delft, The Netherlands, July 15-19, 2024}, D.~Kulic, G.~Venture, K.~E. Bekris, and E.~Coronado, Eds., 2024. [Online]. Available: \url{https://doi.org/10.15607/RSS.2024.XX.120}
\BIBentrySTDinterwordspacing

\bibitem{openxembodiment2024}
A.~O’Neill, A.~Rehman, A.~Maddukuri, A.~Gupta, A.~Padalkar, A.~Lee, A.~Pooley, A.~Gupta, A.~Mandlekar, A.~Jain, \emph{et~al.}, ``Open x-embodiment: Robotic learning datasets and rt-x models: Open x-embodiment collaboration 0,'' in \emph{2024 IEEE International Conference on Robotics and Automation (ICRA)}.\hskip 1em plus 0.5em minus 0.4em\relax IEEE, 2024, pp. 6892--6903.

\bibitem{walke2023bridgev2}
\BIBentryALTinterwordspacing
H.~R. Walke, K.~Black, T.~Z. Zhao, Q.~Vuong, C.~Zheng, P.~Hansen{-}Estruch, A.~W. He, V.~Myers, M.~J. Kim, M.~Du, A.~Lee, K.~Fang, C.~Finn, and S.~Levine, ``Bridgedata {V2:} {A} dataset for robot learning at scale,'' in \emph{Conference on Robot Learning, CoRL 2023, 6-9 November 2023, Atlanta, GA, {USA}}, ser. Proceedings of Machine Learning Research, J.~Tan, M.~Toussaint, and K.~Darvish, Eds., vol. 229.\hskip 1em plus 0.5em minus 0.4em\relax {PMLR}, 2023, pp. 1723--1736. [Online]. Available: \url{https://proceedings.mlr.press/v229/walke23a.html}
\BIBentrySTDinterwordspacing

\bibitem{jiang2025dexmimicgen}
Z.~Jiang, Y.~Xie, K.~Lin, Z.~Xu, W.~Wan, A.~Mandlekar, L.~J. Fan, and Y.~Zhu, ``Dexmimicgen: Automated data generation for bimanual dexterous manipulation via imitation learning,'' in \emph{2025 IEEE International Conference on Robotics and Automation (ICRA)}.\hskip 1em plus 0.5em minus 0.4em\relax IEEE, 2025, pp. 16\,923--16\,930.

\bibitem{bjorck2025gr00t}
J.~Bjorck, F.~Casta{\~n}eda, N.~Cherniadev, X.~Da, R.~Ding, L.~Fan, Y.~Fang, D.~Fox, F.~Hu, S.~Huang, \emph{et~al.}, ``Gr00t n1: An open foundation model for generalist humanoid robots,'' \emph{arXiv preprint arXiv:2503.14734}, 2025.

\bibitem{Pi2025pi05}
P.~Intelligence \emph{et~al.}, ``$\pi_{0.5}$: a vision-language-action model with open-world generalization,'' 2025.

\bibitem{shi2025hirobot}
L.~X. Shi, B.~Ichter, M.~Equi, L.~Ke, K.~Pertsch, Q.~Vuong, J.~Tanner, A.~Walling, H.~Wang, N.~Fusai, \emph{et~al.}, ``Hi robot: Open-ended instruction following with hierarchical vision-language-action models,'' \emph{arXiv preprint arXiv:2502.19417}, 2025.

\bibitem{GRT2025geminiRobotics}
{Gemini Robotics Team and others}, ``Gemini robotics: Bringing {AI} into the physical world,'' 2025.

\bibitem{zawalski2024ecot}
M.~Zawalski, W.~Chen, K.~Pertsch, O.~Mees, C.~Finn, and S.~Levine, ``Robotic control via embodied chain-of-thought reasoning,'' 2024.

\bibitem{chen2025ecot-lite}
W.~Chen, S.~Belkhale, S.~Mirchandani, O.~Mees, D.~Driess, K.~Pertsch, and S.~Levine, ``Training strategies for efficient embodied reasoning,'' 2025.

\bibitem{lin2025onetwovla}
\BIBentryALTinterwordspacing
F.~Lin, R.~Nai, Y.~Hu, J.~You, J.~Zhao, and Y.~Gao, ``Onetwovla: {A} unified vision-language-action model with adaptive reasoning,'' \emph{CoRR}, vol. abs/2505.11917, 2025. [Online]. Available: \url{https://doi.org/10.48550/arXiv.2505.11917}
\BIBentrySTDinterwordspacing

\bibitem{huang2022inner}
W.~Huang, F.~Xia, T.~Xiao, H.~Chan, J.~Liang, P.~Florence, A.~Zeng, J.~Tompson, I.~Mordatch, Y.~Chebotar, \emph{et~al.}, ``Inner monologue: Embodied reasoning through planning with language models,'' \emph{arXiv preprint arXiv:2207.05608}, 2022.

\bibitem{zeng2022socratic}
A.~Zeng, M.~Attarian, B.~Ichter, K.~Choromanski, A.~Wong, S.~Welker, F.~Tombari, A.~Purohit, M.~Ryoo, V.~Sindhwani, J.~Lee, V.~Vanhoucke, and P.~Florence, ``Socratic models: Composing zero-shot multimodal reasoning with language,'' 2022.

\bibitem{ichter2022saycan}
\BIBentryALTinterwordspacing
B.~Ichter, A.~Brohan, \emph{et~al.}, ``Do as {I} can, not as {I} say: Grounding language in robotic affordances,'' in \emph{Conference on Robot Learning, CoRL 2022, 14-18 December 2022, Auckland, New Zealand}, ser. Proceedings of Machine Learning Research, K.~Liu, D.~Kulic, and J.~Ichnowski, Eds., vol. 205.\hskip 1em plus 0.5em minus 0.4em\relax {PMLR}, 2022, pp. 287--318. [Online]. Available: \url{https://proceedings.mlr.press/v205/ichter23a.html}
\BIBentrySTDinterwordspacing

\bibitem{nakamoto2024steering}
M.~Nakamoto, O.~Mees, A.~Kumar, and S.~Levine, ``Steering your generalists: Improving robotic foundation models via value guidance,'' \emph{arXiv preprint arXiv:2410.13816}, 2024.

\bibitem{ma2025valuefxn}
\BIBentryALTinterwordspacing
Y.~J. Ma, J.~Hejna, C.~Fu, D.~Shah, J.~Liang, Z.~Xu, S.~Kirmani, P.~Xu, D.~Driess, T.~Xiao, O.~Bastani, D.~Jayaraman, W.~Yu, T.~Zhang, D.~Sadigh, and F.~Xia, ``Vision language models are in-context value learners,'' in \emph{The Thirteenth International Conference on Learning Representations, {ICLR} 2025, Singapore, April 24-28, 2025}.\hskip 1em plus 0.5em minus 0.4em\relax OpenReview.net, 2025. [Online]. Available: \url{https://openreview.net/forum?id=friHAl5ofG}
\BIBentrySTDinterwordspacing

\bibitem{huang2023voxposer}
W.~Huang, C.~Wang, R.~Zhang, Y.~Li, J.~Wu, and L.~Fei-Fei, ``{Voxposer: Composable 3D Value Maps for Robotic Manipulation with Language Models},'' \emph{arXiv preprint arXiv:2307.05973}, 2023.

\bibitem{liang2023code}
J.~Liang, W.~Huang, F.~Xia, P.~Xu, K.~Hausman, B.~Ichter, P.~Florence, and A.~Zeng, ``{Code as Policies: Language Model Programs for Embodied Control},'' in \emph{IEEE International Conference on Robotics and Automation}.\hskip 1em plus 0.5em minus 0.4em\relax IEEE, 2023, pp. 9493--9500.

\bibitem{fangandliu2024moka}
K.~Fang, F.~Liu, P.~Abbeel, and S.~Levine, ``{MOKA: Open-World Robotic Manipulation through Mark-Based Visual Prompting},'' \emph{Robotics: Science and Systems (RSS)}, 2024.

\bibitem{du2023guiding}
Y.~Du, O.~Watkins, Z.~Wang, C.~Colas, T.~Darrell, P.~Abbeel, A.~Gupta, and J.~Andreas, ``Guiding pretraining in reinforcement learning with large language models,'' 2023.

\bibitem{wang2023voyager}
G.~Wang, Y.~Xie, Y.~Jiang, A.~Mandlekar, C.~Xiao, Y.~Zhu, L.~Fan, and A.~Anandkumar, ``Voyager: An open-ended embodied agent with large language models,'' \emph{arXiv preprint arXiv:2305.16291}, 2023.

\bibitem{bhat2024grounding}
V.~Bhat, A.~U. Kaypak, P.~Krishnamurthy, R.~Karri, and F.~Khorrami, ``Grounding llms for robot task planning using closed-loop state feedback,'' \emph{arXiv preprint arXiv:2402.08546}, 2024.

\bibitem{chen2022open}
B.~Chen, F.~Xia, B.~Ichter, K.~Rao, K.~Gopalakrishnan, M.~S. Ryoo, A.~Stone, and D.~Kappler, ``Open-vocabulary queryable scene representations for real world planning,'' \emph{arXiv preprint arXiv:2209.09874}, 2022.

\bibitem{smith2025steer}
L.~Smith, A.~Irpan, M.~G. Arenas, S.~Kirmani, D.~Kalashnikov, D.~Shah, and T.~Xiao, ``Steer: Flexible robotic manipulation via dense language grounding,'' in \emph{2025 IEEE International Conference on Robotics and Automation (ICRA)}.\hskip 1em plus 0.5em minus 0.4em\relax IEEE, 2025, pp. 16\,517--16\,524.

\bibitem{openai2025gpt5}
OpenAI \emph{et~al.}, ``{GPT-5 System Card},'' \emph{OpenAI Blog}, 2025.

\bibitem{zhao2023chunking}
\BIBentryALTinterwordspacing
T.~Z. Zhao, V.~Kumar, S.~Levine, and C.~Finn, ``Learning fine-grained bimanual manipulation with low-cost hardware,'' in \emph{Robotics: Science and Systems XIX, Daegu, Republic of Korea, July 10-14, 2023}, K.~E. Bekris, K.~Hauser, S.~L. Herbert, and J.~Yu, Eds., 2023. [Online]. Available: \url{https://doi.org/10.15607/RSS.2023.XIX.016}
\BIBentrySTDinterwordspacing

\bibitem{vosylius2024instant}
V.~Vosylius and E.~Johns, ``Instant policy: In-context imitation learning via graph diffusion,'' \emph{arXiv preprint arXiv:2411.12633}, 2024.

\bibitem{fu2024context}
L.~Fu, H.~Huang, G.~Datta, L.~Y. Chen, W.~C.-H. Panitch, F.~Liu, H.~Li, and K.~Goldberg, ``In-context imitation learning via next-token prediction,'' \emph{arXiv preprint arXiv:2408.15980}, 2024.

\bibitem{lewis2020retrieval}
P.~Lewis, E.~Perez, A.~Piktus, F.~Petroni, V.~Karpukhin, N.~Goyal, H.~K{\"u}ttler, M.~Lewis, W.-t. Yih, T.~Rockt{\"a}schel, \emph{et~al.}, ``{Retrieval-augmented generation for knowledge-intensive NLP tasks},'' \emph{Advances in neural information processing systems}, vol.~33, pp. 9459--9474, 2020.

\bibitem{glossop2025cast}
C.~Glossop, W.~Chen, A.~Bhorkar, D.~Shah, and S.~Levine, ``Cast: Counterfactual labels improve instruction following in vision-language-action models,'' \emph{arXiv preprint arXiv:2508.13446}, 2025.

\end{thebibliography}
